\pdfoutput=1
\documentclass[11pt]{article}
\usepackage{emnlp2021}
\usepackage{times}
\usepackage{latexsym}
\usepackage[T1]{fontenc}
\usepackage[utf8]{inputenc}
\usepackage{microtype}
\usepackage{graphicx}
\usepackage{multirow}
\usepackage{xcolor}
\usepackage{amsmath,amssymb,stackengine}
\usepackage{array}

\usepackage{svg}

\author{%
Daniela Brook Weiss$^{\,1}$%
~\;~\;~ Paul Roit$^{\,1}$%
~\;~\;~ Ayal Klein$^{\,1}$%
~\;~\;~ Ori Ernst$^{\,1}$%
{\bf~\;~\;~ Ido Dagan$^{1}$}\\
  $^{1}$Computer Science Department, Bar-Ilan University\\%
  \texttt{\footnotesize{\{dani.b.weiss,plroit,ayal.s.klein,oriern\}@gmail.com}}\, \texttt{\footnotesize{dagan@cs.biu.ac.il}}
}

\title{QA-Align: Representing Cross-Text Content Overlap \\ by Aligning Question-Answer Propositions}

\date{}
\begin{document}
\maketitle

\begin{abstract}
Multi-text applications, such as multi-document summarization, are typically required to model redundancies across related texts.    
Current methods confronting
consolidation struggle to fuse overlapping information.
In order to explicitly represent content overlap, we propose to align predicate-argument relations across texts, providing a potential scaffold for information consolidation. 
We go beyond clustering coreferring  
mentions, and instead model overlap with respect to redundancy at a propositional level, rather than merely detecting shared referents. Our setting exploits QA-SRL, utilizing question-answer pairs to capture predicate-argument relations, facilitating laymen annotation of cross-text alignments.
We employ crowd-workers for constructing a dataset of QA-based alignments,
and present a baseline QA alignment model trained over our dataset.
Analyses show that our new task is semantically challenging, capturing content overlap
beyond lexical similarity and complements cross-document coreference with proposition-level links, offering potential use for downstream tasks.
\end{abstract}

\section{Introduction}
\label{sec:intro}

End-to-end neural methods have become the de-facto standard for natural language understanding models.
While these often work well for single document tasks, tasks that consider multiple textual inputs remain more challenging.  
A key difficulty concerns consolidating information from different, possibly redundant texts, which is crucial for applications such as multi-document summarization (MDS), sentence fusion \cite{mckeown-2010-3rd, thadani-mckeown-2013-supervised-fusion-dataset-4th} or multi-hop question-answering \cite{welbl2018multihop-reasoning, feldman2019multi-hop-QA}. 
Previous works show that MDS methods for example, 
often just concatenate rather than merge inputs \cite{lebanoff-etal-2019-analy-fusion}, or erroneously consolidate on non-coreferring elements \cite{lebanoff-etal-2020-poc}. Recognizing such challenges, a few earlier MDS works \cite{liu-etal-2015-amr-first, liao-etal-2018-amr-second, shapira2017interactive} attempted at leveraging semantic structures for consolidation, such as
Abstract Meaning Representation \cite{banarescu2013AMR} or Open Knowledge Graph \cite{wities2017consolidatedOKR}, however, these initial means were found either too fine-grained (AMR) or immature (OKR) for efficient downstream consolidation.

\begin{figure}[t]
  \centering
  \includegraphics[width=\columnwidth]{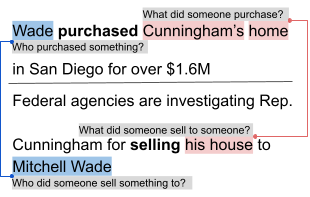}
  \caption{An example of two alignments expressing the same propositions captured by QASRL question-answers.}
  \label{fig:align_qasrl_ex}
\end{figure}

In this paper, we propose that a useful step toward effective consolidation is to detect and align \emph{minimal} propositions
 that refer to the same information.
Therefore, we choose to capture propositions at the fine-grained level of  predicate-argument relations,
each representing a single predication involving two sentence elements, along with the semantic role that relates them. Identifying these types of alignments at the propositional level would facilitate recognizing information redundancies and salience. Aligning propositions across documents may also prove useful to a variety of cross-text tasks, such as knowledge-base population, Multi-hop QA and cross document event extraction.

Consider the example in \autoref{fig:align_qasrl_ex}; our alignments represent that both sentences express similar propositional information --- sentence A talks about a ``buying'' event while sentence B is framing the same event through the ``seller'' perspective, essentially capturing the same reversed roles.
In previous predicate-argument alignment works \cite{roth-frank--first-task-2012-aligning-predicate, wolfe-etal-2013-parma}, predicates and arguments were aligned individually, rather than aligning the predicate argument relation as a whole, along with its semantic role.
Our approach can be seen as a conceptual extension over the established Coreference Resolution framework, while handling content overlap at the more complex level of semantic relations (conveying stated information) rather than entity and event mentions (denoting referents). This difference in alignment scope also pertains to the cross-document coreference dataset used in these prior works, as we analyze and compare to in \S \ref{sec:ecb-comparison}.

Unlike earlier work \citep{roth-frank--first-task-2012-aligning-predicate} that leveraged structured SRL to align predicate-argument-structures,
we leverage the Question-Answer driven Semantic Role Labeling paradigm (QA-SRL) \cite{He2015qasrl}.
QA-SRL captures predicate-argument relations using a naturally phrased question-answer pair, where the question type (who, what, where, etc.) reflects the role being captured, while the answer denotes the argument. 
For example, in \autoref{fig:align_qasrl_ex} sentence A, the Agent role (A0) of the predicate ``purchase'' is captured through the question-answer \textit{Who \underline{purchased} something? --- \underline{Wade}}.
Once such relations are captured by QAs, we can align these QAs to capture similar propositional information expressed in a pair of sentences (see \autoref{fig:align_qasrl_ex}).
Since QAs are naturally intelligible and do not require a pre-defined schema, they can be easily annotated and used by non-experts.  

Our contributions are outlined as follows: We introduce the QA-Align task, modeling cross-document proposition-level content overlap as QA-based alignments (\S \ref{sec:task_def}). 
We show that this task, though semantically challenging, is attainable via crowd annotations, and publish our guidelines and crowdsourcing methodology. 
In addition, we compile and release our crowdsourced QA-Align dataset (accompanied by new QA-SRL annotations), over semantically similar paired texts (\S \ref{corpus_creation}) that we collect. \footnote{Our Code and data can be found here: \url{https://github.com/DanielaBWeiss/QA-ALIGN}}
We further analyze the quality of our data and compare it to an established cross-document (CD) coreference benchmark, ECB+ \cite{ecb+} (\S \ref{sec:data-analysis}). Finally, we implement a baseline modeling approach for QA-Align (\S \ref{sec:qa_parser}), and present an analysis of appealing potential downstream use, over a sentence fusion task (\S \ref{sec:extrinsic}). 
\section{Related Work}
\label{sec:background}
Proposition alignment is closely related to the Cross Document Coreference Resolution (CDCR) task \cite{gooi2004cross-doc-coref-CDCR, mayfield2009cross-doc-CR}. 
This task concerns clustering together entity or event mentions across topically related documents that refer to the same ``real world'' element, (entity or event).
It has drawn substantial recent attention \cite{barhom2019coref, zeng-etal-2020-event-coref, cattan2020coref, yu-yin-roth2020paired}, as its considered a fundamental intermediate task for cross-text language understanding. 
While the concept of coreference is essential for defining our alignment criteria (\S \ref{sec:task_def}), we consider matching predicate--argument relations as expressing alignments of propositional information, and accordingly as a useful component in capturing information correspondence between texts, unlike prior approaches which match individual event and entity mentions disjointly.

Our work is largely inspired by the earlier predicate-argument alignment task \cite{roth-frank--first-task-2012-aligning-predicate,roth-frank-2015-inducing,wolfe-etal-2013-parma,wolfe-etal-second-2015-predicate}. 
These works used either a semantic role labeler to align predicates to predicates \cite{roth-frank--first-task-2012-aligning-predicate}, or annotated coreference chains to align both predicates and arguments, though disjointly evaluated \cite{wolfe-etal-2013-parma}. 
Unlike our work, these works did not take into consideration the relationship between the predicates and arguments, as expressed by the semantic role relating them.
In this sense, these prior works are closer to the CDCR setting, whereas we directly align on propositional relations as the core units of interest, which, as we show later, captures more exhaustively information overlap across related texts (\S \ref{sec:ecb-comparison}).

As potential uses for downstream tasks, various prior works presume that effectively handling multiple text sources, such as for multi-document summarization (MDS), requires intermediate explicit representations for cross-text content consolidation \cite{liao-etal-2018-amr-second, wities2017consolidatedOKR}.  
Further, modeling redundant or overlapping text has been tackled through the sentence fusion task \cite{barzilay-mckeown-2005-intro-task, marsi-krahmer-2005-explorations-2nd,mckeown-2010-3rd,thadani-mckeown-2013-supervised-fusion-dataset-4th}, and recently as ``disparate'' sentence fusion, targeting related sentences in a single document \cite{nayeem-etal-2018-abstractive,geva2019discofuse,lebanoff-etal-2019-analy-fusion,lebanoff-etal-2019-scoring-pairs,lebanoff-etal-2020-poc}.
In a recent series of investigations,
\citet{lebanoff-etal-2019-analy-fusion} 
highlight sentence fusion as a necessary step for improving summarization. 
In particular, pairs of sentences to be fused were empirically shown as a better source for generating summary sentences than single sentences \cite{lebanoff-etal-2019-scoring-pairs}.
In Section \ref{sec:extrinsic}, we analyze the potential utility of our QA alignments as a redundancy signal in a sentence fusion model. 
\paragraph {QA-SRL}
QA-SRL has been shown to attain high quality annotation of predicate-argument structure for verbs, via crowdsourcing \cite{fitz2018qasrl}, achieving above 90\% coverage of PropBank arguments \cite{roit2020qasrl-gs}. 
In addition, \citet{roit2020qasrl-gs} showed that it captures much implicit information that is often missed by traditional SRL schemes.
For these reasons, we employ the QA-SRL scheme for crowdsourcing propositional alignments and then modeling them in our QA-Align parser (\S \ref{sec:qa_parser}).
\section{Task Definition}
\label{sec:task_def}
Setting the task of aligning semantic content across text requires defining the informational units that are to be aligned. 
Moving from coreference resolution onto the propositional level, it is desired to pinpoint the minimal unit of propositional information, at which information overlap may be flexibly identified.
Inspired by the seminal Neo-Davidsonian approach \cite{parsons1990neo-davidsonian} and following previous works \cite{roth-frank--first-task-2012-aligning-predicate,roth-frank-2015-inducing, wolfe-etal-2013-parma, liu-etal-2015-amr-first}, we view a single predicate-argument relation as an atomic propositional unit. 
As mentioned, we capture these using QA-SRL \cite{He2015qasrl} Question-Answer (QA) pairs, as was shown in \autoref{fig:align_qasrl_ex}.

The current scope of our work considers alignments between two sentences.
We define the task as follows: given a pair of texts concerning (partly) overlapping scenarios, along with their predicate-argument structure representation --- consisting of QA-SRL QAs in our setting, we want to find all cross-text alignments between QAs that refer to the same fact within the context of their sentences.

More concretely, our annotation guidelines, targeted for trained crowd workers (see \S \ref{corpus_creation}), require aligned QAs to correspond on the following elements:
\begin{enumerate}
    \item The main verbs in both questions refer to the same event.
    \item Answers refer to the same referent or meaning. 
    \item Questions convey the same relation between the two, i.e.\ ask about same role of information regarding the predicate.
\end{enumerate}
\autoref{fig:align_qasrl_ex} is an example of our crowd-sourced QA-alignments.
In this example, our crowd-workers intuitively aligned the corresponding two roles for the predicates ``buy'' and ``sell'', even though their syntactic structure is reversed, as it is clear that all verbs, answers and questions correspond.

The vast majority of the alignments we observe are 1-to-1, that is, aligning a single QA from each sentence. 
However, we also allow for many-to-many alignments in our annotation protocol,
which constitute about 4\% of our data. These  are required, for instance, to handle light-verb constructions, as in the following example:
\begin{enumerate}
    \item \textit{The owner hopes to \textbf{display} the painting}
    \item \textit{He hopes the Picasso painting will \textbf{go on display} tomorrow}
\end{enumerate}
where the QA \textit{What might someone display? --- the painting} is aligned to the set of two QAs \textit{What will something go on? --- display}, and \textit{What will go on something? --- the Picasso painting}.
Such alignments must also be minimal, meaning that taking out any QA from the alignment would posit it invalid.

Leveraging verbal QA-SRL, our task formulation targets verbal predicates, leaving the coverage of other predicate types, including nominalizations \cite{klein2020qanom}, for future work.

\begin{table*}[hbt!]
\resizebox{\textwidth}{!}{%
\begin{tabular}{l|p{12cm}}
Sentence Contributor & The  \textbf{alleged bomb-builder was arrested in Egypt.}\\ \hline
Sentence Contributor & ABC News reported that  \textbf{the alleged bomb maker behind the London attacks was arrested in Egypt.}\\ \hline
SCU Label & Bomb-maker arrested in Egypt
\end{tabular}%
}
\caption{Source summary sentences with their SCU contributing spans (in bold), and their given SCU Label.}
\label{tab:scu_ex}
\end{table*}
\begin{table}[t!]
\centering
\resizebox{\columnwidth}{!}{%
\begin{tabular}{llll}
          & Train & Dev   & Test \\ \hline
Num. Paired Texts        & \textbf{1374} & \textbf{400}  & \textbf{605}  \\ \hline
Avg Num. Alignments   & 2.3           & 3.0           & 2.5           \\ \hline
Num. QASRL questions & 19191*  & 4299  & 8048  \\ \hline
Total alignments & \textbf{3162} & \textbf{1205} & \textbf{1508} \\ \hline
No alignments  & 29\% & 16\% & 24\% \\ \hline
Many-to-many  & 3\% & 4\% & 4\%           
\end{tabular}% 
}
\caption{Statistics on the dataset collected. *Questions are produced using the QASRL parser. Over 94\% of our many-to-many alignments are 2-to-1.}
\label{tab:dataset-stats}
\end{table}
\section{Dataset}
\label{corpus_creation}

\subsection{Data Sources}
Annotating cross-text propositional alignments requires a collection of semantically related sentences.
Previous works \cite{roth-frank--first-task-2012-aligning-predicate, wolfe-etal-2013-parma, lebanoff-etal-2019-scoring-pairs} collected such sentence pairs based on computed similarity scores, which biases the data toward sentences that the utilized model already recognizes as similar. 
To avoid such bias, we leverage available human annotations that identified information overlap from various sources. 
Such sources yield naturally occurring texts, representative for multi-text applications, while challenging models with realistic lexical and content diversity. We next describe the concrete sources for our dataset.

\paragraph{ECB+}
The Event Coreference Bank (ECB+, an extension over ECB) \cite{ecb+} provides annotations for coreferring event and entity mentions across a ``topic" --- a set of documents related to the same event. 
We use this source to collect related pairs of sentences that share at least one coreferring verbal event mention.
To control for data diversity, we take at most 6 sentence-pairs from each ECB+ topic, those with varying degrees of similarity based on a shared number of coreferring mentions.

\paragraph{DUC}
The Document Understanding Conference
(DUC)\footnote{https://www-nlpir.nist.gov/projects/duc/data.html, years used 2005-2008} and the Text Analysis Conference (TAC)\footnote{https://tac.nist.gov/, years used 2009-2011} both provide  multi-document summarization evaluation datasets. 
In particular, we leveraged available manual Pyramid annotations over these datasets \cite{nenkova-passonneau-2004-pyramid}.
Under this scheme, annotators extract from reference summaries small units of information, termed ``summary content units'' (SCUs), and then match them with corresponding information in system summaries to evaluate their quality.
SCUs that repeat in several reference summaries create a cluster of similar \textit{SCU contributors}, as can be seen in Table \ref{tab:scu_ex} in bold, and are given a manually written SCU label describing the main content of the cluster.

Inspired by \citet{thadani-mckeown-2013-supervised-fusion-dataset-4th}, we collect clusters of summary sentences that include matching SCU contributors. 
Then, we take each pair of summary sentences that appear in the same cluster to be a sentence-pair instance for our dataset.

\begin{table}[ht]
\centering
\resizebox{\columnwidth}{!}{%
\begin{tabular}{l|ccc}
 & \textbf{ECB+} & \textbf{DUC} & \textbf{MN} \\ \hline
Paired Sentences & 611 & 1028 & 740 \\
Unique Sentences & 704 & 1563 & 1265 \\
Topics & 86 & 347 & 503 \\
Avg ROUGE 2 between pairs& 0.21 & 0.12 & 0.13 
\end{tabular}%
}
\caption{Distribution of our crowdsourced train, dev, and test data from the three different multi-text sources.}
\label{tab:dataset-sources}
\end{table}

\paragraph{MultiNews}

MultiNews (MN) \cite{fabbri-etal-2019-multi} is a recent Multi Document Summarization dataset, containing clusters of news articles along with human-written summaries.
Recently, \citet{superpal-multinews} crowdsourced alignments between semantically matching proposition spans, of a similar nature to Pyramid SCUs, across documents and summaries. 
To generate our data, we collected from their gold data pairs of sentences that include aligned spans.

Table \ref{tab:dataset-sources} details the source distributions of our data. 
The average ROUGE-2 similarity score across paired sentences indicates that we indeed achieve a dataset of semantically similar sentences, yet exhibiting a limited degree of lexical overlap, providing challenging alignment cases. We maintained original train/dev/test splits for each corpora where suitably available and created our own otherwise. For more details regarding our dataset creation, see Appendix \ref{app:dataset_construction_further}.

\subsection{Crowdsourcing}
Aiming at high quality annotation, we applied a controlled crowdsourcing methodology \cite{roit2020qasrl-gs}, over the Amazon Mechanical Turk platform.
Crowd workers were selected and trained, ensuring a reliable, scalable, cheap and rapid process.

\subsubsection{QA-SRL Annotation}
We begin with crowdsourcing the prerequisite QA-SRL annotation over our data. 
We followed the guidelines and protocol of \citet{roit2020qasrl-gs}, 
utilizing their released annotation tools.
For our dev and test sets, we collected a single worker's annotation per predicate. Evaluating their performance against an expert, we found its quality  to match that reported in \citet{roit2020qasrl-gs}.
For the larger training set, we used the available QA-SRL parser\footnote{\url{http://github.com/nafitzgerald/nrl-qasrl}} \cite{fitz2018qasrl} to produce question-answer pairs, thus 
reducing costs and demonstrating the feasibility of automatically generating question-answer pairs to be subsequently aligned by our proposed approach.\footnote{A quality account for the automatically produced QAs and subsequent alignments can be found in Appendix \ref{app:qasrl-parser-assessment}.}

\subsubsection{Annotating QA-Alignments}
Following the controlled crowdsourcing methodology, we first publish a ``trap'' task, for the purpose of identifying an initial pool of workers that perform adequately on the QA alignment task. 
We then begin a training process, where workers read detailed annotation guidelines and answer a follow-up survey.\footnote{Our task interfaces and our detailed guidelines can be found in Appendix \ref{app:crowdsourcing} and at the end of the Appendix respectively.} 
Next, they go through 3 practice rounds, during which they receive detailed feedback.
Subsequently, we selected 6 of the best performing workers for dataset annotation. 

For our training set, each sentence pair was annotated by a single worker, who detected all QA alignments for that pair. 
For the development and test sets, each pair was annotated independently by two workers (\textit{aligners}), yielding two QA alignment sets. A third worker then arbitrated any disagreements, yielding the final set of QA alignments for the sentence pair.
We qualitatively observed that most disagreements arise from recall misses, 
while the rest arise from true semantic ambiguity.
Section \ref{sec:IAA} presents inter-annotator agreement analysis, for both individual workers as well as for the arbitrated annotations.

Table \ref{tab:dataset-stats} presents the details of our final dataset. As shown, the average number of QA alignments per sentence pair in the arbitrated Dev and Test sets ranges from 2.5 to 3. As each alignment corresponds to a matched (propositional) predicate-argument relation, this indicates a substantial level of information overlap captured in our data.
We also note that our training set contains a high percentage of ``no alignments'', meaning instances where workers deemed nothing to align. 
This fact is likely attributed to using the QA-SRL parser to produce the QA pairs for each sentence (versus using human QA-SRL annotations), which is known to struggle with coverage \cite{fitz2018qasrl}.\footnote{For more examples of our crowdsourced QA-alignments, see Appendix \ref{annotation_examples}.}

\paragraph{Compensation} 
Initially, workers are compensated for the time and effort invested during their training process.
In data production, in addition to a fixed compensation per annotation instance (i.e.\ sentence pair), we use bonuses to account for instance complexity, which is measured by the size of QA sets for both sentences.  
Overall, aligners receive on average \textbf{\$12}/hour, or  \textbf{30\textcent} per instance, while arbitrators are paid \textbf{22\textcent} per instance, given it is an easier task. To conclude, a final instance in our gold dataset costs \textbf{82\textcent} (two \textit{aligners} and one \textit{arbitrator}), while a ``silver'' instance in our training set costs \textbf{30\textcent}. 
% REMOVE FOR ISCOL, BRING BACK LATER
 \begin{figure*}[t]
   \centering
   \includegraphics[width=\textwidth]{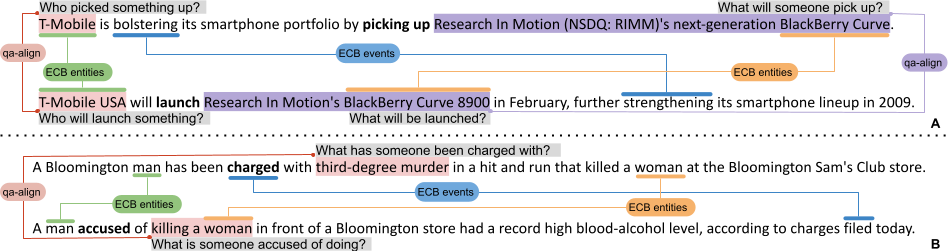}
   \caption{Examples of comparing ECB+ coreference annotations to QA-alignments. \textbf{A. Missed ECB+ event alignment} ECB+ fails to align the main event (in \textbf{bold}) while correctly aligning its core participants. In comparison, QA-Align annotates the arguments using propositions that depict the main event. QA-Align also aligns more propositions that include other ECB+ corefering events (blue underline), not shown for brevity.  \textbf{B. Argument out of scope for ECB+} QA-Align aligns the theme of the charged/accused event mentions (in pink), while such type of participants are out of scope for ECB+, which considers only entities as arguments.}
   \label{fig:final_comparison_to_ecb}
 \end{figure*}
\section{Dataset Quality Evaluation}
\label{sec:data-analysis}

\subsection{Inter-Annotator Agreement}
\label{sec:IAA}
To estimate the reliability of our annotated alignments, we measure inter-annotator agreement (IAA), both between single workers as well as between ``teams" that produce our final annotation (i.e.\ two aligners and an arbitrator).
Agreement between two alignment sets for an input sentence pair is measured simply by an exact match criterion. A QA-alignment is regarded as true positive if it fully matches a reference QA-alignment, while measuring agreement by the yielded F1 score.

The mean agreement between two single annotators (the two \textit{aligners}) for the sentence pairs in our dataset is \textbf{71} F1.
In 44\% of the instances, the two aligners fully agreed in their annotation, where the remaining 56\% instances are passed to a third worker for arbitration.
Hence, since final annotations are collected using a team of workers, evaluating the overall consistency of our data requires assessing team-vs-team agreement. 
For this evaluation, we performed 5 experiments, each including two teams (disjoint worker triplets) which annotate the same 20 sentence pairs (for a total of 100 instances over the 5 experiments). 
Averaging the agreement score across the 5 experiments results in \textbf{84} F1, expressing a substantial data consistency for a challenging semantic task.

Notably, the arbitration procedure reduced disagreements by almost 50\%, and indicated that a major part of worker-vs-worker disagreements are not a matter of controversy. Rather, they stem from the difficulty of a single annotator to exhaustively identify all valid alignments between the two sets of QAs, justifying our annotation protocol. 

\subsection{Comparison to ECB+ Coreference Annotations}
\label{sec:ecb-comparison}
In this section, we assess and analyze the additional scope of information correspondence captured by our proposition-level annotation scheme, relative to the common paradigm in prior work which builds on event and entity coreference. 
To this end, we leverage the prominent ECB+ cross-document coreference dataset as a reference point over the corresponding subset of our data.

To compare our QA-alignments dataset with ECB+, we automatically induce one-to-one proposition level \textit{ECB-based alignments} from event and entity mentions.
We re-use gold QA-SRL data as the underlying propositions, and consider two QA pairs as aligned if their predicates corefer as mentions of the same event, and if their answers have any overlap with coreferring entity mentions.
This procedure assumes that, in most cases, a match of both the question predicates and their answers would correspond to a match of the two QA pairs as a whole, including a match of their semantic role and compositional meaning.

We found that \textbf{70.3}\% of the ECB-based alignments are covered by our QA-alignments. 
To understand the gap, we manually analyzed a random sample of 66 out of the 302 ECB-based alignments not covered by the QA-alignments in our dataset.
The majority of these (77\%) were erroneous or redundant, due to artifacts in their automatic generation process.\footnote{Redundancies commonly derive from within-document entity coreference, where two mentions (e.g. \textit{the man} and \textit{he}) yield two redundant QAs (\textit{Who came? --- the man} and \textit{Who came? --- he}).
Erroneous ECB-based alignments mostly consist of QAs whose answers do not corefer, but do encompass coreferring entity mentions, e.g. the misleading \textbf{he} in “When did someone come? -- after \textbf{he} finished working”.}
20\% of the non-matched ECB-based alignments correspond to proposition alignments that were effectively captured in our data, using different expressions in the sentence
(see Example 4 in Table \ref{tab:appendix_example_table} in Appendix \ref{annotation_examples}, where a QA-alignment captures more information than an ECB+ coreference).
Importantly, only two alignments (3\%) reflect true misses of our annotations, evaluating the interpolated effective recall of our data relative to ECB+ at \textbf{99\%}.

On the other hand, \textbf{37.3}\% of our gold QA-alignments are not covered by the ECB-based alignments.
Manually inspecting these illuminates the additional scope gained from addressing information overlap at the level of predicate-argument relations, as captured by QA pairs, rather than at the level of individual event and entity mentions.
First, entity coreference often falls short in accounting for corresponding propositions, since many verbal arguments (answers) --- e.g.\ those comprised of full clauses --- are not entity mentions. In \autoref{fig:final_comparison_to_ecb}-\textbf{B}, for example, although ECB+ aligns the events \textit{charged / filed}, their corresponding ``Theme'' arguments (highlighted in pink) cannot be aligned using entity coreference.
More generally, identifying overlapping information may fail using naive coreference criteria.
Consider the \textit{picking up} / \textit{launch} events in \autoref{fig:final_comparison_to_ecb}-\textbf{A}.  The two verbs are quite distinct in terms of their basic lexical meaning, which is probably why ECB+ refrains from annotating them as coreferring events. 
Nevertheless, as manifested by the predicate-argument relations involving these mentions, it is clear that in the given context, these two event mentions convey the same information.

All in all, these analyses show that proposition-level alignments are required in order to properly capture information overlap, complementing the more elementary task of traditional event and entity coreference resolution. 
\section{Baseline QA-Align Model}
\label{sec:qa_parser}

\begin{table*}[ht]
\centering
\resizebox{\textwidth}{!}{%
\begin{tabular}{llll}
\multicolumn{4}{l}{\textbf{Example QA-Align Model Input}} \\
Input A & \multicolumn{3}{l}{\begin{tabular}[c]{@{}l@{}}Who did someone [P] fire [/P] ? [Q] The Philadelphia 76ers [P] fired [/P] [A] coach Maurice Cheeks [/A] on Saturday, \\ one day after the team continued its slide with a season-worst offensive effort, dpa reported.\end{tabular}} \\\hline
Input B & \multicolumn{3}{l}{\begin{tabular}[c]{@{}l@{}}Who was [P] fired [/P] ? [Q] If you don't know by now: you disappoint in the NBA, you get canned. Today,  \\ \relax
[A] Maurice Cheeks [/A] became the fifth coach [P] fired [/P] within the first quarter of the season.\end{tabular}}
\end{tabular}%
}
\caption{Example input encoding for our baseline models (\S \ref{sec:qa_parser_description}).} 
%Question and context are concatenated and denoted with a separator special added token, along with additional markup tokens denoting the predicates and argument answers.}
\label{tab:model-input}
\end{table*}

\subsection{Model Description}
\label{sec:qa_parser_description}
Taking a simplistic modeling approach as an initial baseline, we reduce the QA alignment prediction problem into a binary classification task. 
Let $(S_1, S_2)$ be a sentence pair, where each sentence is provided with a set $R$ of QA-SRL relations (QAs): 
$$ r_i^s  \in R_s  \: , \;  1 \le i \le | R_s | \: , \; s \in \{1,2\}  $$

Our baseline considers only 1:1 alignments (recall this covers 96\% of the cases in our data). 
Given a single pair of QAs $(r_i^1, r_j^2)$  as a candidate alignment along with their contexts, each comprising of the sentence itself and its predecessor sentence, the model predicts the probability that the pair of QAs is aligned.
We serialize the full information of a candidate QA pair into the input sequence, and feed it into a pre-trained language model with a binary classifier on top, as typical for sequence classification.
An example input for our model can be seen in Table \ref{tab:model-input}.
For each QA, we concatenate the question and the context sentences, while denoting the predicate and the answer span with special markup tokens, attuning the model toward the correspondences within and across candidates \cite{baldini-soares-etal-2019-matching,lebanoff-etal-2020-learning}.
More details about the training procedure appear in Appendix \ref{qa-model-train}.
During inference, we treat each QA-to-QA candidate alignment as an edge and weigh it with the probability score produced by the model, filtering edges scored under a threshold $\tau = 0.5$. 
The final alignment set is then decoded using maximal bipartite matching.

We experimented with BERT \cite{devlin2019bert}, RoBERTa \cite{liu2019roberta}, and CorefRoBERTa \cite{ye-etal-2020-corefBERT}\footnote{The paper presented CorefBERT, while we use a released CorefRoBERTa version.} as the pretrained models.
Following previous works on CD coreference \cite{ecb+, barhom2019coref, zeng-etal-2020-event-coref} and predicate-argument alignment \cite{wolfe-etal-2013-parma, roth-frank--first-task-2012-aligning-predicate}, we compare our models to a lemma-based 
baseline method.
Similarly to the alignment criterion applied for the ECB-based analysis of our crowdsourced data (\S \ref{sec:ecb-comparison}), 
our lemma baseline model aligns QA pairs in which the two predicates, as well as the head words of the answer spans, share a lemma. 

\subsection{Model Performance}
      
\begin{table}[t]
\centering
\resizebox{\columnwidth}{!}{%
\begin{tabular}{l|ccc|ccc}
\multirow{2}{*}{Method / Model} & \multicolumn{3}{c|}{Dev} & \multicolumn{3}{c}{Test} \\
 & P & R & F1 & P & R & F1 \\ \hline
Lemma & 89 & 35 & 50 & \textbf{89} & 30 & 45 \\
Bert-base-cased & 72 & 51 & 60 & 62 & 41 & 49   \\
Roberta-base  & 66 & 50 & 57 & 58 & 44 & 50  \\
CorefRoberta & 71 &59 &64 & 60 & \textbf{48} & \textbf{53} \\
\end{tabular}
}
\caption{Precision, Recall, and F1 results for multiple QA-Align baselines.}
\label{tab:model-stats}

\end{table}

\label{model-performance}
Results are shown in Table \ref{tab:model-stats}. \footnote{Skipped many-to-many alignments are counted as recall misses.}
Notably, the lemma baseline performance is relatively low,  reinforcing that aligning predicate-argument propositions across texts is challenging.\footnote{
As a rough non-comparable reference point, it is interesting to note that the ``same lemma'' baseline for CD event coreference achieves 76.5 CoNLL F1 on ECB+, providing a strong baseline which recent state-of-the-art models only surpass by 4 to 8 F1 points.  \cite{cattan2020coref, zeng-etal-2020-event-coref}.}
Applying pre-trained language models yields a modest yet clear improvement. The best performance is obtained using CorefRoBERTa~\cite{ye-etal-2020-corefBERT}, which is specialized on coreferential reasoning by further pre-training on predicting masked entity mentions repeating in a single document.
Overall, we suggest that QA alignment provides a challenging task for future research, with much room for improvement, particularly considering the high inter-annotator agreement of our laymen annotators (\S \ref{sec:IAA}). 

Analysis shows that the lemma baseline, while exhibiting high precision (as expected), misses many alignments where the predicates or arguments are paraphrastic rather than lexically identical.
On the other hand, we find that the alignment model often misses alignments of similar questions that include different ``Wh'' words.
For instance, the QA alignment containing \textit{Why was someone pulled over?} and \textit{What was someone pulled over for?} is missed by our model. 
Indeed, recognizing ``same-role'' paraphrastic questions is a known challenge in the context of QA-SRL evaluation \cite{fitz2018qasrl, roit2020qasrl-gs}, with potential value also for QA-SRL downstream tasks, and is left for future work. 
\section{Analysis of Potential Downstream Use}
\label{sec:extrinsic}

\begin{table*}[ht!]
\centering
\resizebox{\textwidth}{!}{%
\begin{tabular}{ll}
\textbf{Example Fuse-Align Input} & \begin{tabular}[c]{@{}l@{}}{\color{blue}Law enforcement agencies} [P1] {\color{blue}use} [\textbackslash P1] [A1] {\color{blue}dogs} [\textbackslash A1] worldwide. <\slash s> \\
{\color{orange}Dogs perform many different law-enforcement tasks around the world.} <\slash s> \\
City and county police agencies, customs departments, fire departments, the Secret Service, \\
highway patrol, border patrol, military bases and some prisons in the US and many \\
other countries [P1] {\color{blue}use} [\textbackslash P1] [A1] {\color{blue}dogs} [\textbackslash A1] {\color{blue}to help in law enforcement work.}\end{tabular} \\ \hline
\textbf{Baseline Output} & {\color{orange}Dogs perform many different law-enforcement tasks around the world} \\
\textbf{Fuse-Align output} & \begin{tabular}[c]{@{}l@{}}{\color{blue}Law enforcement agencies use dogs to help in law enforcement}\end{tabular}
\end{tabular}%
}
\caption{An example input for the Fuse-Align model marked with predicted alignments (the baseline does not include these tokens). The indices in the special tokens indicate which spans are aligned across sentences, <\slash s> are sentence separators. The orange span is the one contributing to the baseline's output, while the blue spans contribute to Fuse-Align's output, showcasing a merge across redundant and complementary information that our alignment model identifies.
}
\label{tab:fusion-ex}
\end{table*}

Aiming at an initial analysis of the potential extrinsic utility of our alignments, we experiment with using them as a signal for the sentence fusion task
\cite{barzilay-mckeown-2005-intro-task,marsi-krahmer-2005-explorations-2nd}.
Sentence fusion is a summarizing task applied for the single sentence level, thus providing a suitable use-case for exploring the extrinsic utility of our sentence-level alignments.
Indeed, it was shown that fusion models tend to generate naive extractive  summaries that are copied from a single source \cite{lebanoff-etal-2019-analy-fusion}, rather than merging information from multiple sources.
In that vein, we aim to examine whether explicitly incorporating alignments into the fusion task would encourage a model to attend to corresponding information originated in multiple sentences.

\subsection{The Sentence Fusion Experiment}

\citet{barzilay-mckeown-2005-intro-task} formulated sentence fusion as a generation task that consolidates the information across multiple input sentences into a single sentence.
Here, we adopt and reproduce the fusion dataset constructed by \citet{thadani-mckeown-2013-supervised-fusion-dataset-4th} (referred in \S \ref{corpus_creation}), which is the most recent source for \textit{multi-document} sentence fusion (vs. the different type of fusion within a single document; see Appendix \ref{app:fusion-data} for more details regarding the dataset).
In this dataset, the input for fusion instances consists of clusters of two to four source sentences, originating in different documents, which need to be summarized.

We create a new modern baseline for the dataset of \citet{thadani-mckeown-2013-supervised-fusion-dataset-4th}, which outperforms their pre-neural one, evaluated using bigram-F1.
As a baseline end-to-end fusion model, we employ the pre-trained auto-encoder BART \cite{lewis2020bart}, which has achieved state-of-the-art results on summarization tasks.
In comparison, we predict QA alignments for the fusion data using our best reported model, and then incorporate them into the fusion model (termed Fuse-Align) using an input-augmentation technique (resembling the one in \S \ref{sec:qa_parser_description}).
As shown in Table \ref{tab:fusion-ex}, the input encompasses the alignment information by attaching indexed markup tokens around aligned predicates and arguments.\footnote{In this initial trial, we do not incorporate the questions in our alignments.}

\subsection{Results and Analysis}
We create 20 different variations of the dataset by randomly shuffling the order of the input sentences in each input, and take the average ROUGE 2 (R2) across all runs.
We find that the baseline and Fuse-Align models achieve a similar R2 of 41 and 40 respectively. 
Although ROUGE scores are similar, upon closer examination we find the outputs of both models to be of different nature.

In order to quantify these differences, we automatically classify fusion outputs stemming from single versus multiple source sentences.
Based on lexical overlap, we link every word in the fused output to its contributor source sentences, if exist. 
An output is considered multi-sourced or \textit{consolidating} if it consists of different words exclusively originating from separate sources (at least two).
This classification is related to the distinction of abstractive vs.\ extractive summarization, where merging multiple sources in a summary sentence falls under the abstractive summarization regime, and is not possible in extractive summarization.

Based on this classification, we find that Fuse-Align significantly creates substantially more consolidating outputs (\textbf{30}\%) compared to the baseline model (\textbf{20}\%), as illustrated in Table \ref{tab:fusion-ex} (see Appendix \ref{app:fusion_output_ex} for more examples).\footnote{In both models the majority is still single-sourced, probably due to the ``extractive'' nature of the this data \cite{thadani-mckeown-2013-supervised-fusion-dataset-4th}.}

To further investigate qualitative differences, we analyze a sample of 50 multi-source outputs from each model. 
We find that both models consistently produce grammatical and faithful outputs for these short input/output pairs, and exhibit a similar level of compression. 
The key difference attributed to our Fuse-Align input augmentation is thus encouraging the BART seq-to-seq model to merge multi-source information 50\% more frequently than over the raw input, in the baseline configuration, and also triggering merges of more than two input sentences, which never happens for the baseline.

To conclude, though further investigation of this trend is worthwhile, this analysis demonstrates a promising potential 
% and need for further investigation 
utilization of propositional alignments.
The fact that adding alignment information to BART's input encourages it merge information sources much more frequently is notable, especially as sequence-to-sequence MDS models often tend to produce mostly extractive outputs \cite{lebanoff-etal-2019-analy-fusion}.
\section{Conclusion}
We present a new task for aligning propositional predicate-argument information, captured via the question-answer pairs of QA-SRL. We compile and publish a QA-Align dataset and present our crowdsourcing methodology, showing that a high inter-annotator agreement is achievable for this challenging semantic task even for non-expert annotators. We compare our annotation scheme to that of ECB+ with respect to aligning information across related texts, and show that alignments at the level of predicate-argument relations capture substantially more information correspondences than aligning individual referents.
In addition, we present a baseline QA-Align model, 
and utilize its predicted alignments to increase information consolidation within a high-redundancy sentence fusion task.
Our exploratory findings warrant further future investigations concerning the potential of predicate-argument alignments for modeling cross-document consolidation.  

\section*{Acknowledgments}
We would like to thank the anonymous reviewers for their thorough and insightful comments.
The work described herein was supported in part by grants from Intel Labs, Facebook, and the Israel Science Foundation grant 1951/17.
\bibliography{9_qaalign.bib}
\bibliographystyle{acl_natbib}

%%% For Camera-Ready, comment out to integrate appendices into the same PDF file as main paper:
\appendix
\section{Examples of Crowdsourced QA-Alignments}
\newcolumntype{L}[1]{>{\raggedright\let\newline\\\arraybackslash\hspace{0pt}}m{#1}}
\begin{table*}[ht]
\centering
\resizebox{\textwidth}{!}{%
\begin{tabular}{llll}
\multirow{5}{*}{\textbf{Ex 1}} & \multicolumn{3}{l}{Sentence A: An earthquake measuring 4 . 6 \textbf{rattled} Sonoma and Lake counties early Thursday, according to the U.S. Geological Survey.} \\
 & \multicolumn{3}{l}{Sentence B: The temblor \textbf{struck} about 26 miles north of Santa Rosa in the Geysers area.} \\ \cline{2-4} 
 & \textbf{Aligned QAs from A} &  & \textbf{Aligned QAs from B} \\
 & What \textbf{rattled} something? --- An earthquake measuring 4 . 6 & --Align1-- & Who \textbf{struck} somewhere? --- The temblor \\
 & \begin{tabular}[c]{@{}l@{}}\begin{tabular}[c]{@{}l@{}}What did something \textbf{rattle}? --- Lake counties\\ What did something \textbf{rattle}? --- Sonoma\end{tabular}\end{tabular} &  --Align2-- & \begin{tabular}[c]{@{}l@{}}\begin{tabular}[c]{@{}l@{}}Where did someone \textbf{strike}? --- about 26 \\ miles north of Santa Rosa \\ in the Geysers area\end{tabular}\end{tabular} \\ \cline{2-4} 
 & \multicolumn{3}{l}{} \\
\multirow{5}{*}{\textbf{Ex 2}} & \multicolumn{3}{l}{Sentence A: A jury in eastern Oklahoma has \textbf{convicted} a 27 - year - old man for killing  his girlfriend.} \\
 & \multicolumn{3}{l}{Sentence B: Man \textbf{found} GUILTY of shooting girlfriend} \\ \cline{2-4} 
 & \textbf{Aligned QAs from A} &  & \textbf{Aligned QAs from B} \\
 & Who did someone \textbf{convict}? --- a 27 - year - old man &  --Align1-- & Who was \textbf{found} as something? --- Man \\
 &  &  & What was someone \textbf{found} as? --- GUILTY \\ \cline{2-4} 
 & \multicolumn{3}{l}{} \\
\multirow{4}{*}{\textbf{Ex 3}} & \multicolumn{3}{l}{Sentence A: Piccard \textbf{got} Chinese airspace clearance before starting but was limited to a narrow strip, forbidden north of the 26th parallel.} \\
 & \multicolumn{3}{l}{Sentence B: They gained vital time by \textbf{obtaining} flight permission from China before taking off.} \\ \cline{2-4} 
 & \textbf{Aligned QAs from A} &  & \textbf{Aligned QAs from B} \\
 & What did someone \textbf{get}? --- airspace clearance & --Align1-- & What did someone \textbf{obtain} from someone? --- flight permission \\ \cline{2-4} 
 & \multicolumn{3}{l}{} \\
\multirow{5}{*}{\textbf{Ex 4}} & \multicolumn{3}{l}{Sentence A: Woman Killed In Queens Hit - And - Run , Driver \textbf{Charged}} \\
 & \multicolumn{3}{l}{\begin{tabular}[c]{@{}l@{}}Sentence B: An allegedly intoxicated driver who tried to flee after striking and fatally injuring a woman in Queens has been \textbf{charged} \\ in her murder, according to police.\end{tabular}} \\ \cline{2-4} 
 & \textbf{Aligned QAs from A} &  & \textbf{Aligned QAs from B } \\
 & Who was \textbf{charged} with something? --- Driver &  --Align1-- & Who was \textbf{charged} in something? --- An allegedly intoxicated driver \\
 & Why was someone \textbf{charged} with something? --- Woman Killed &  --Align2-- & \begin{tabular}[c]{@{}l@{}}\begin{tabular}[c]{@{}l@{}}Why was someone \textbf{charged }in something? --- tried to flee after \\ striking and fatally injuring a woman in Queens\end{tabular}\end{tabular}
\end{tabular}%
}
\caption{A table showcasing our crowdsourced alignments.}
\label{tab:appendix_example_table}
\end{table*}

\label{annotation_examples}
Table \ref{tab:appendix_example_table} presents a few examples collected by our crowd-workers following our crowdsourcing methodology.
Examples 1 \& 3 present an example of lexically different entities being captured and aligned using our proposition alignments.
Example 2 is a light-verb construction case talked about in \S\ref{sec:task_def}.
Example 4 (Align2) presents an instance where QA-SRL captures a \textit{causality} discourse relation (in both sentences), not explicitly manifested in syntactic structure (and indeed missed by an SRL parser trained on PropBank).  
This enables our workers to align semantically rich propositions that go beyond syntactic analysis. 
Consequently, workers align on these relations rather then the more narrow alignment captured by the ECB-based alignments to sentence B's QA: \textit{What was someone charged in? --- her murder}, resulting from a coreference link between \textit{Woman} and \textit{her}.

\begin{table}[]
\centering
\resizebox{\columnwidth}{!}{%
\begin{tabular}{lllll}
\multirow{4}{*}{Num. Instances} &  & \textbf{Train} & \textbf{Dev} & \textbf{Test} \\
 & \textbf{ECB+} & 355 & 116 & 140 \\
 & \textbf{DUC} & 534 & 182 & 312 \\
 & \textbf{MN} & 485 & 102 & 153
\end{tabular}%
}
\caption{Distribution of our train/dev/test splits. Using Official splits of ECB+ and MN, while splitting our DUC in accordance with \citet{thadani-mckeown-2013-supervised-fusion-dataset-4th}'s fusion dataset split, in order not to compromise our training when testing on our extrinsic task.}

\label{app:sources-splits}
\end{table}

\section{Dataset Construction -- Further Details}
\label{app:dataset_construction_further}
In this section we further explain the details regarding our dataset creation corpus.
Given the smaller number of topics annotated in ECB+ and DUC, we don't want to take too many pairs of sentences that overlap over the same content, therefore we filter out any pair of sentences with ROUGE 2 above 0.9. In addition, we attach previous sentences to each instance for context, used when presenting the instance to crowd-workers, and in our QA-Alignment model.
In terms of train, dev, and test splits, we use official splits where available, and create our own where needed. For ECB+, we take the official split published; for MultiNews, we also use the official split provided, and in addition, supplement with 5 extra topics from the train and add it to the dev and test (each gets unique 5 topics), given a very small amount of sentence pairs were released with gold annotations (312 and 345 for dev and test respectively). For DUC, an official split is not available, therefore we follow  
\citet{thadani-mckeown-2013-supervised-fusion-dataset-4th}
official sentence fusion data split, based on DUC and TAC's published years. Table \ref{app:sources-splits} presents the sources distributions by training splits. 

\subsection{ECB+}
Due to the relative small number of topics annotated in ECB+ \cite{ecb+}, 
with usually only a handful of shared events per topic and on average only 2 annotated sentences per document, the resulting pairs are not well varied in content and are highly repetitive.
Because of this, we rank the sentence pairs in ECB+ by their number of shared coreferring mentions, and limit to top 6 pairs per topic. In addition, we collect sentence pairs from the lowest levels of shared coreferring mentions, in order to maintain a lexical variety and negative examples in our data.
We select only verbal predicates from ECB+ when creating semantically related pairs.

\subsection{DUC}
In addition to the summary sentence pairs we collect using their shared SCU label, we also collect document to document, and document to summary pairs of sentences using the SCU Marked Corpus \cite{scu-marked-corpus},
which links SCU labels to document source sentences. We don't perform any filtering on DUC, given it's varied in content and lexicality due to the large amount of topics and years available.

\subsection{MultiNews}
For MultiNews (MN), we also extract document to document aligned sentences, using a shared aligned summary sentence as an anchor. Because two different document sentences can be aligned to the same summary sentence, but with different non-overlapping spans, we also filter out any document pairs that have a span IOU (intersection over union) of lower than 0.1. We measure this IOU based on how much overlap the two document sentences have in their aligned spans to the summary sentence they share.

\section{Crowdsourcing}
\label{app:crowdsourcing}
\subsection{User-Interface}
\begin{figure*}[]
    \centering
    \includegraphics[width=\textwidth]{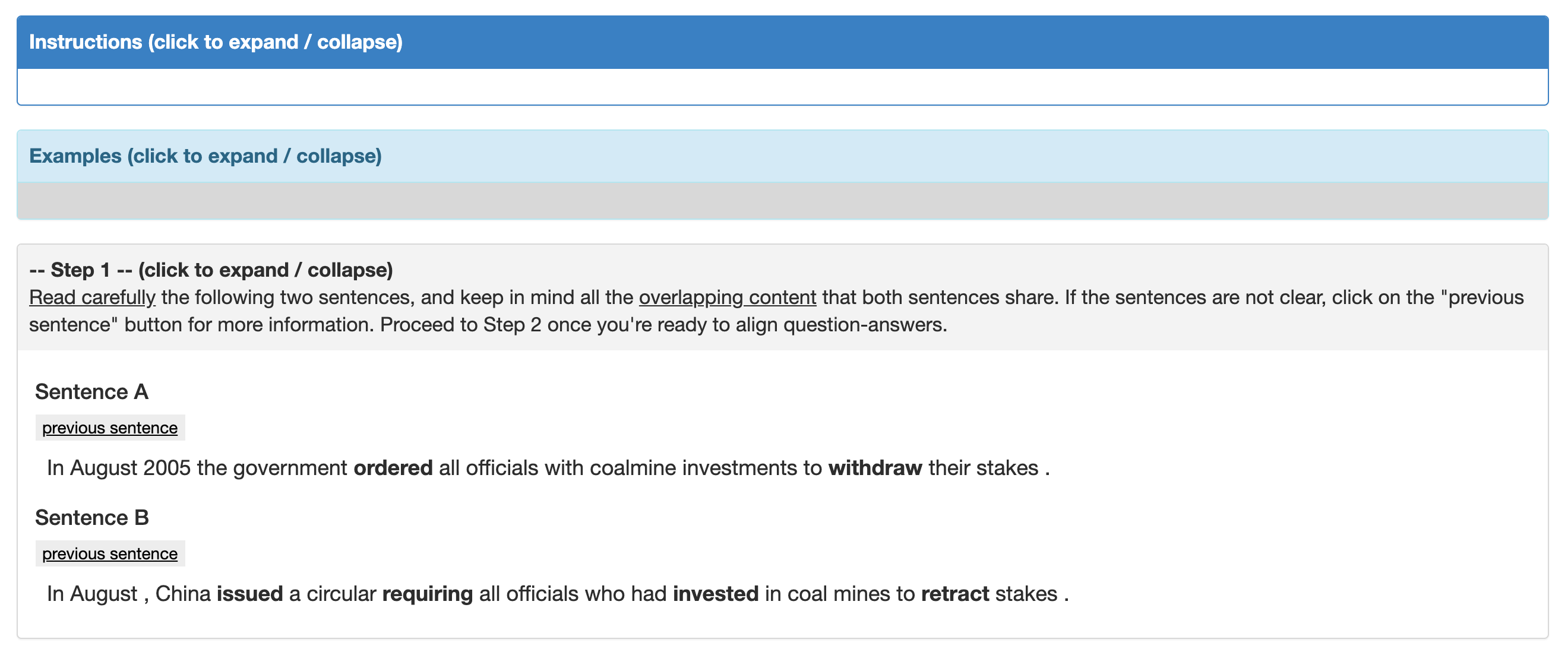}
    \captionof{figure}{QA-Alignment Task Interface Step 1}
    \label{fig:task_ui}
    \includegraphics[width=\textwidth]{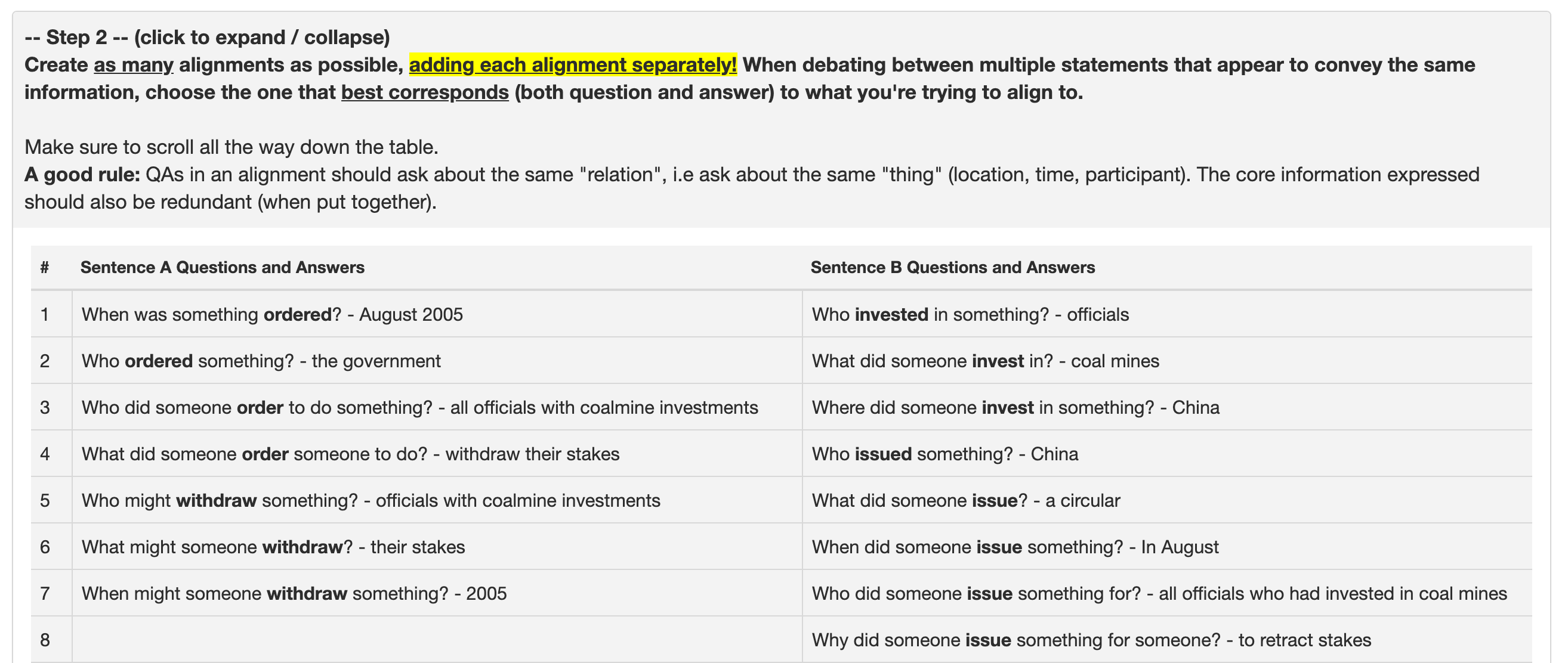}
    \captionof{figure}{QA-Alignment Task Interface Step 2}
    \label{fig:task_ui_2}
\end{figure*}

Figures \ref{fig:task_ui} and \ref{fig:task_ui_2} exhibit Step 1 in our crowdsourcing task on Amazon Mechanical Turk. Workers are instructed to read a pair of sentences carefully, noting any overlap of information they find, and in Step 2 they carefully select QAs from the presented table and create the alignments.

\section{Quality Assessment of QAs Predicted by QASRL Parser}
\label{app:qasrl-parser-assessment}

We make use of an updated version of the QASRL parser published by \citet{fitz2018qasrl},
which according to developer's comment in the repository, reaches 85 F1 on span detection and 47.6\% question prediction accuracy. 
Nevertheless, accurately evaluating produced question is difficult, since various questions can validly describe the same semantic role, depending on the predicate's lexical semantics and on context. This results in a under-estimation of question quality by the question prediction accuracy measure.
A manual assessment of the QASRL parser based on an improved evaluation protocol\cite{roit2020qasrl-gs} 
suggest that its role precision is relatively good (83.9), but its role recall is mediocre (64.3), as has been previously acknowledged \cite{fitz2018qasrl}.
Nevertheless, we opt for having a more sparse training set while showing it is possible for future endeavors to scale up training set annotation using an off-the-shelve QASRL parser.

Limited QA coverage might also result in lower QA alignment quality. 
In order to assess the quality of alignments based on parser produced QAs, we give one worker two sets of the same 30 instances, once with the parser-generated QAs and once with gold collected QAs. 
We find that the worker achieved  \textbf{79} F1 agreement with herself, whereas the average number of alignments went from 2.3 with gold QAs to 1.9 with predicted QAs. 
To conclude, using the parser results in a modest alignment coverage decrease, while saving the costs of collecting gold QASRL annotations.

\section{QA-Alignment Model Training}
\label{qa-model-train}
For training, we apply negative sampling with a 1/10 positive/negative candidate alignment ratio, compared to an approximate 1/50 naturalistic ratio.
Specifically, we sample negative alignments that include verbs or answers contained in the positive set, which we find leading to better performance, presumably because it leaves the ``harder'' cases for the model to distinguish from.
As for training parameters, we use a learning rate of 3e-5 for both BERT and CorefRoBERTa, and 2e-5 for RoBERTa; for all we use the Adam optimizer with an epsilon of 1e-8, and max length tokens of 256, in addition to adding the special tokens used in our input to the model's vocabulary. We finetune BERT and CorefRoBERTa for 5 epochs, and 4 epochs for RoBERTa.

\section{Sentence Fusion Dataset Creation}
\label{app:fusion-data}
For the extrinsic evaluation of our collected alignments on sentence fusion (\S\ref{sec:extrinsic}) we reproduce the dataset constructed by \citet{thadani-mckeown-2013-supervised-fusion-dataset-4th}. 
We note that while more recent datasets for the sentence fusion task exist \cite{geva2019discofuse,lebanoff-etal-2020-poc},
these works concern disparate sentences coming from a single document, which have minor content overlap \cite{lebanoff-etal-2019-analy-fusion}
and thus diverge from the multi-doc setting. 

As mentioned in \S\ref{corpus_creation},
the fusion data is generated using SCU labels from the DUC datasets, where SCU labels function as the target fusion outputs, and the sentence contributors as input sentences.
Although the original dataset is not available, they published detailed instructions that enabled us to recreate the data. We note also that the nature of this fusion dataset is not entirely comparable to the one we extracted from DUC explained in \S\ref{corpus_creation}.
The fusion dataset that was described in 
Thadani and McKeown (2013) 
contained post-processed sentence clusters (which we use as is only for the fusion experiment, we don't do any processing in our QA-Alignment data). 
This was because the authors in the paper applied a number of heuristics on the pairs created by the SCU labels in DUC, in order to remove noisy and longer sentences with small SCU contributors, discarding shorter and maintaining only highly overlapping sentences. This means that although the resulting dataset is highly relevant for cross-text information overlap, it also means that this fusion dataset created might not entirely reflect real document and summary sentences found ``in the wild'', however, it is the only sentence-level fusion dataset currently available.

Although \citet{thadani-mckeown-2013-supervised-fusion-dataset-4th} originally reported 1858 instances, they informed us they were using an unreleased version of DUC 2005. 
Our regenerated dataset thus consists of 1705 fusion instances, distributed into the author's original 70/30 train/test split, using DUC years 2005-2007 for test, TAC 2011 for dev, and TAC 2008-2011 for train.

\section{Fusion Output Examples}
\label{app:fusion_output_ex}
\begin{table*}[ht]
\centering
\resizebox{\textwidth}{!}{%
\begin{tabular}{ll}
\textbf{Fusion Input} & \begin{tabular}[c]{@{}l@{}}  [A2] She [\textbackslash A2] called it " [P5] Change [\textbackslash P5] [A5] Your Life TV [\textbackslash A5] " ... <\slash s> \\ Recently [A4] she [\textbackslash A4] has been [P4] presenting [\textbackslash P4] a " [P5] change [\textbackslash P5] [A5] your life [\textbackslash A5] " theme \\ \relax [P3] focusing [\textbackslash P3] on [A3] who you are [ \textbackslash A3] , and a truth - will - set - you - free attitude , influenced by her \\ own traumatic childhood .  [A1] Her show [\textbackslash A1] {\color{blue}is all about} [P1] {\color{blue}learning} [\textbackslash P1] {\color{blue}who YOU are.} <\slash s> \\\relax [A42] \emph{The
Oprah Winfrey Show} [\textbackslash A42] [P3] {\color{orange}features} [\textbackslash P3] [A3] [P4] {\color{orange}talking} [\textbackslash P4] {\color{orange}cures} [\textbackslash A3] {\color{orange}and} [P12] \\
{\color{orange}learning} [\textbackslash P12] [A1] {\color{orange}who you are} [\textbackslash A1].\end{tabular} \\ \hline
\textbf{Baseline Output} & {\color{orange}The Oprah Winfrey Show features talking cures and learning who you are} \\
\textbf{Fuse-Align Output} & {\color{blue}The Oprah Winfrey Show is all about learning who YOU are} \\
\multicolumn{2}{l}{} \\
\textbf{Fusion Input} & \begin{tabular}[c]{@{}l@{}}{\color{orange}A year later 150,000 were in prefab homes} and only 26,000 still in tent cities. <\slash s> A year later about 26,000 \\{\color{blue} survivors were living in tent cities} that had initially held {\color{blue}some 120,000}; and some 150,00 people were still \\ in prefabricated houses.\end{tabular} \\ \hline
\textbf{Baseline Output} & {\color{orange}A year later 150,000 were in prefab homes} \\
\textbf{Fuse-Align Output} & {\color{blue}Some 150,000 survivors were living in tent cities} \\
\multicolumn{2}{l}{} \\
\textbf{Fusion Input} &
\begin{tabular}[c]{@{}l@{}}{\color{blue}In December 1994,} [A1] Tony Blair [\textbackslash A1], Labour leader, and probable prime minister \\ after an upcoming election, signaled a constitutional battle with the Tories \\
by [P312] {\color{blue}setting} [\textbackslash P312] 
{\color{blue}out} [A2] [A3] {\color{blue}proposals} [\textbackslash A3] {\color{blue}for devolution} for Scotland and Wales [\textbackslash A2] \\ <\slash s> {\color{orange}Labour has committed to the creation of a Welsh assembly}, and [A1] party leader \\ {\color{blue}Tony Blair} [\textbackslash A1] [P12] {\color{blue}set} [\textbackslash P12] {\color{blue}out} [A2] {\color{blue}proposals for devolution} [\textbackslash A2].\end{tabular} \\ \hline
\textbf{Baseline Output} & {\color{orange}Labour has committed to the creation of a Welsh assembly} \\
\textbf{Fuse-Align output} & \begin{tabular}[c]{@{}l@{}}{\color{blue}In December 1994, Tony Blair set out proposals for devolution}\end{tabular}
\end{tabular}%
}
\caption{Example outputs for both fusion models. The input represents what our Fuse-Align models accepts, while the baseline takes the same input just without the special tokens.}
\label{tab:fusion_table_examples}
\end{table*}
In Table \ref{tab:fusion_table_examples} we display further fusion input examples that are fed to our baseline models, marked with special tokens as the predicted alignments.
Fuse-Align's outputs in examples 1 and 3 display a behavior that is more abstractive than the baseline's (as discussed in \S\ref{sec:extrinsic}), 
able to identify corresponding parts across source sentences, and ``fuse'' them into a single output. On the other hand, example 2 is an instance where our model also merges information, however, doing so on the incorrect entities (the 150,000 are not the survivors that live in tent cities, as the model predicts).
Since our QA-Align model did not predict any alignments on this pair, this error could also result from not having any aligned propositions to attend to.

\end{document}